\documentclass[a4paper,11pt, reqno]{amsart}
\usepackage{array}
\usepackage{graphicx}
\usepackage{float}
\usepackage{amsmath,latexsym}
\usepackage{amssymb}
\usepackage{geometry}
\usepackage{amsfonts}
\usepackage{hyperref}
\usepackage{color}
\usepackage{multirow}
\def\R{\mathbb{R}}

\newcommand{\dsum}{\displaystyle\sum}

\usepackage{setspace}
\allowdisplaybreaks

\let\origmaketitle\maketitle
\def\maketitle{
  \begingroup
  \def\uppercasenonmath##1{} 
  \let\MakeUppercase\relax 
  \origmaketitle
  \endgroup
}
\title[]{\large Robust Optimal Classification Trees under Noisy Labels}

\author[V. Blanco, A. Jap\'on \MakeLowercase{and} J. Puerto]{{\large V\'ictor Blanco$^\dagger$, Alberto Jap\'on$^\ddagger$ and  Justo Puerto$^\ddagger$}\medskip\\
$^\dagger$Institute of Mathematics, Universidad de Granada\\
$^\ddagger$Institute of Mathematics, Universidad de Sevilla}

\date{\today}

\begin{document}

\maketitle

\begin{abstract}
In this paper we propose a novel methodology to construct Optimal Classification Trees that takes into account that noisy labels may occur in the training sample. Our approach rests on two main elements: (1) the splitting rules for the classification trees are designed to maximize the separation margin between classes applying the paradigm of SVM; and (2) some of the labels of the training sample are allowed to be changed during the construction of the tree trying to detect the label noise. Both features are considered and integrated together to design the resulting  \textit{Optimal} Classification Tree. We present a Mixed Integer Non Linear Programming formulation for the problem, suitable to be solved using any of the available off-the-shelf solvers. The model is analyzed and tested on a battery of standard datasets taken from UCI Machine Learning repository, showing the effectiveness of our approach.
\end{abstract}

\section{Introduction}

Discrete Optimization has experienced a tremendous growth in the last decades, both in its theoretical and practical sides, partially provoked by the emergence of new computational resources as well as real-world applications that have boosted this growth. This impulse has also motivated the use of Discrete Optimization models to deal with problems involving a large number of variables and constraints, that years before would have not been  possible to be dealt with. One of the fields in which Discrete  Optimization has caused a larger impact is in Machine Learning.  The incorporation of binary decisions to the classical approaches as Support Vector Machine~\cite{svm}, Classification Trees~\cite{cart}, Linear Regression and Clustering, amongst other, has considerably enhanced their performance in terms of accuracy and interpretability (see e.g. \cite{benati2017clustering,bertsimas2017optimal,BJPP20,BJP20,blanco2018locating,blanquero2020selection,blanquero2020cost,drucker1997support,gaudioso2017lagrangian}). In particular, one of the most interesting applications of Machine Learning is that related with Supervised Classification.

Supervised classification aims at finding hidden patterns from a training sample of labeled data in order to predict the labels of out-of-sample data.  Several methods have been proposed in order to construct highly predictive classification tools. Some of the most widely used methodologies are based on Deep Learning mechanisms~\cite{Agarwal_2018}, $k$-Nearest Neighborhoods \cite{knn1,knn}, Na\"ive Bayes \cite{nb}, Classification Trees \cite{cart,friedman2001} and Support Vector Machines~\cite{svm}. The massive use of these tools has induced, in many situations, that malicious adversaries adaptively manipulate their data to mislead the outcome of an automatic analysis, and new classification rules must be designed to handle the possibility of this noise in the training labels. A natural example are the spam filters for emails, where malicious emails are becoming more difficult to automatically be detected since they have started to incorporate patterns that typically appear in legitimate emails. As a consequence, the development of robust methods against these kind of problems has attracted the attention of researchers (see e.g.,  \cite{bertsimas2019robust,BJP2020b}).

In the context of binary classification problems, Support Vector Machines (SVM), introduced by Cortes and Vapnik~\cite{svm},  builds the decision rule by means of a separating hyperplane with large margin between classes. This hyperplane is obtained by solving a Non Linear Problem (NLP), in which the goal is to separate data by their two differentiated classes, maximizing the margin between them and minimizing the misclassification errors. In Classification and Regression Trees (CART), firstly introduced by Breiman et. al \cite{cart}, one constructs the decision rule based on a hierarchical relation among a set of nodes which is used to define paths that lead observations from the root node (highest node in the hierarchical relation), to some of the leaves in which a class is assigned to the data. These paths are obtained according to different optimization criteria over the predictor variables of the training sample. The decision rule comes up naturally, the classes predicted for new observations are the ones assigned to the terminal nodes in which observations fall in. Historically, CART is obtained heuristically through a greedy approach, in which each level of the tree is sequentially constructed: Starting at the root node and using the whole training sample, the method  minimizes an impurity measure function obtaining as a result a split that divides the sample into two disjoint sets which determine the two descendant nodes. This process is repeated until a given termination criterion is reached (minimum number of observations belonging to a leaf, maximum depth of the tree, or minimum percentage of  observations of the same class on a leaf, among others). In this approach, the tree  grows following a top-down greedy approach, an idea that is also shared in other popular decision tree methods like C4.5~\cite{quinlan93} or ID3~\cite{quinlan96}. The advantage of these methods is that the decision rule can be obtained rather quickly even for large training samples, since the whole process relies on  solving manageable problems at each node. Furthermore, these rules are interpretable since the splits only take into account information about lower or upper bounds on a single feature. Nevertheless, there are some remarkable disadvantages in these heuristic methodologies. The first one is that they may not obtain the \textit{optimal} classification tree, since they look for the best split locally at each node, not taking into account the splits that will come afterwards. Thus, these local branches may not capture the proper structure of the data, leading to misclassification errors in out-of-sample observations. The second one is that, specially under some termination criteria, the solutions provided by these methods can result into very deep (complex) trees, resulting in overfitting and, at times, loosing interpretability of the classification rule. This difficulty is usually overcome by pruning the tree as it is being constructed by comparing the gain on the impurity measure reduction with  respect to the complexity cost of the tree.

Mixing together the powerful features of standard classification methods and Discrete Optimization has motivated the study of supervised classification methods under a new paradigm (see \cite{bertsimas2019machine}). In particular, recently, Bertsimas and Dunn \cite{bertsimas2017optimal} introduced the notion of \textit{Optimal Classification Trees} (OCT) by approaching CART under optimization lens, providing a Mixed Integer Linear Programming formulation to optimally construct Classification Trees. In this formulation, binary variables are introduced to model the different decisions to be taken in the construction of the trees: deciding whether a split is applied and if an observation belongs to a terminal node. Moreover, the authors proved that this model can be solved for reasonable size datasets, and equally important, that for many different real datasets, significant improvements in accuracy with respect to CART can be obtained. In contrast to the standard CART approach, OCT builds the tree by solving a single optimization problem taking into account (in the objective function) the complexity of the tree,  avoiding post pruning processes. Moreover, every split is directly applied in order to minimize the misclassification errors on the terminal nodes, and hence, OCTs are more likely to capture the essence of the data. Furthermore, OCTs can be easily adapted in the so-called OCT-H model to decide on splits based on hyperplanes (oblique) instead of in single variables. Another remarkable advantage of using optimization tools in supervised classification methods is that features such as sparsity or robustness, can be incorporated to the models by means of binary variables and constraints~\cite{gunluk2018optimal}. The interested reader is refereed to the recent survey  \cite{OCTsurvey}. We would like to finish this discussion pointing out one of the main differences between SVM and Classification Trees: SVM accounts for misclassification errors based on distances (to the separating hyperplane), i.e., the closer to the correct side of the separating hyperplane, the better, whereas in Classification Trees all misclassified observations are equally penalized.

Recently, Blanco et. al~\cite{BJP2020b} proposed different SVM-based methods that provide robust classifiers under the hypothesis of label noises. The main idea supporting those methods is that labels are not reliable, and in the process of building classification rules it may be beneficial to \textit{flip} some of the labels of the training sample to obtain more accurate classifiers. With this paradigm, one of the proposed methods, RE-SVM, is based on constructing a SVM separating hyperplane, but simultaneously allowing observations to be relabeled during the training process. The results obtained by this method, in datasets in which noise was added to the training labels, showed that this strategy outperforms, in terms of accuracy, classical SVM and other SVM-based robust methodologies. See \cite{bertsimas2019robust} for alternative robust classifiers under label noise.

In this paper we propose a novel binary supervised classification method, called Optimal Classification Tree with Support Vector Machines (OCTSVM), that profits both from the ideas of SVM and OCT to build classification rules. Specifically, our method  uses the hierarchical structure of OCTs, which leads to easily interpretable rules, but splits are based on SVM hyperplanes, maximizing the margin between the two classes at each node of the tree. The fact that the combination of SVM and classification tree tools provides enhanced classifiers is not new. A similar approach can be found in \cite{bennett1998support}. Nevertheless, in that paper the authors analyze the greedy CART strategy by incorporating, sequentially the maximization of the margin, over known assignments of observations to the leaves of the tree. Opposite to that,  OCTSVM does not assume those assumptions and it performs an exact optimization approach. Moreover, this new method  also incorporates decisions on relabeling observations in the training dataset, making it specially suitable for datasets where adversary attacks are suspected. The results of our experiments show that OCTSVM outperforms other existing methods under similar testing conditions.

The rest of the paper is organized as follows. In Section \ref{sec:2} we recall the main ingredients of our approach, in particular, SVM, RE-SVM and OCTs, as well as the notation used through the paper. Section \ref{sec:3} is devoted to introduce our methodology, and presents a valid Mixed Integer Non Linear Progamming (MINLP) formulation. In Section \ref{sec:4} we report the results obtained in our computational experience, in particular, the comparison of our method with OCT, OCT-H and the greedy CART. Finally, some conclusions and further research on the topic are drawn in Section \ref{sec:5}.

\section{Preliminaries}\label{sec:2}

In this section we recall the main ingredients in the approach that will be presented in Section \ref{sec:3} which allows us to construct robust classifiers under label noises, namely, Support Vector Machines with Relabeling (RE-SVM) and Optimal Classification Trees with oblique splits (OCT-H).

All through this paper, we consider that we are given a training sample $\mathcal{X} = \left\lbrace (x_1,y_1)\right.$, $\left.\ldots (x_n,y_n),  \right\rbrace \subseteq \R^p \times \left\lbrace -1, +1 \right\rbrace$, in which $p$ features have been measured for a set of $n$ individuals ($x_1, \ldots, x_n$) as well as a $\pm 1$ label is also known for each of them ($y_1, \ldots, y_n$). The goal of supervised classification is, to derive, from $\mathcal{X}$, a decision rule $D_\mathcal{X}: \R^p \rightarrow \{-1,1\}$ capable to accurately predict the right label of out-sample observations given only the values of the features. We assume, without loss of generality that the features are normalized, i.e., $x_i \in [0,1]^p$.

One of the most used optimization-based method to construct classifiers for binary labels is SVM~\cite{svm}. This classifier is constructed by means of a separating hyperplane in the feature space, $\mathcal{H} = \left\lbrace z \in\mathbb{R}^p : \omega'z + \omega_0 = 0 \right\rbrace$, such that the decision rule becomes:
$$
D_\mathcal{X}^{SVM} (x) = \left\{\begin{array}{cl}
-1 & \mbox{if $\omega'z + \omega_0\leq 0$},\\
1 & \mbox{if $\omega'z + \omega_0\geq 0$}.
\end{array}\right.
$$
To construct such a hyperplane, SVM chooses the one that simultaneously maximizes the separation between classes and minimizes the errors of misclassified observations. These errors are measures proportional to the distances (in the feature space) from the observations to their label half-space. SVM can be formulated as a convex non linear programming (NLP) problem. This approach allows one for the use of a kernel function as a way of embedding the original data in a higher dimension space where the separation may be easier without increasing the difficulty of solving  the problem (the so-called kernel trick).

On the other hand, SVM has also been studied in the context of robust classification. In \cite{BJP2020b} three new models derived from SVM are developed to be applicable to datasets in which the observations may have wrong labels in the training sample. This characteristic is incorporated into the models by allowing some of the labels to be swapped (relabelled) at the same time that the separating hyperplane is built.  Two of these models combine SVM with cluster techniques in a single optimization problem while the third method, the so-called RE-SVM, relabels observations based on misclassification errors without using clustering techniques, what makes it easier to train. The RE-SVM problem can be formulated as:
\begin{align}
\min & \; \frac{1}{2}\|\omega\|_2^2+  c_1 \dsum_{i=1}^n  e_{i}+  c_2\dsum_{i=1}^n\xi_i \label{model:1}\tag{${\rm RE-SVM}$}\\
\mbox{s.t. }\ & (1- 2\xi_i) y_i (\omega'x_i + \omega_0) \geq 1-e_i, &\forall i = 1 , \ldots , n,\label{c2}\\
& \omega \in \mathbb{R}^p, \ \omega_0 \in \mathbb{R}, & \nonumber\\\
&e_i \in \mathbb{R}^+, \ \xi_i \in \left\lbrace 0, 1 \right\rbrace,  & \forall i = 1 , \ldots , n, \nonumber
\end{align}
where $\xi_i$ takes value $1$ if the $i$th observation of the training sample is relabelled, and $0$ otherwise and $e_i$ is the misclassifying error defined as the hinge loss:
 $$
 e_i = \left\{\begin{array}{cl}
\max\{0, 1 - y_i(\omega' x_i + \omega_0)\} & \mbox{if observation $i$ is not relabelled}\\
\max\{0, 1 + y_i(\omega' x_i + \omega_0)\} & \mbox{if observation $i$ is relabelled}
\end{array}\right.,
$$
for $i=1, \ldots, n$. The costs parameters $c_1$ and $c_2$ (unit cost per misclassifying error and per relabelled observation) allows one to find a trade-off between large separation between classes: $c_1$ and $c_2$ are parameters modelling the unit cost of misclassified errors and relabelling, respectively. ($\|\cdot\|_2$ stands for the Euclidean norm in $\R^p$.) Constraints \eqref{c2} assures the correct definition of the hinge loss and relabelling variables.

The problem above can be reformulated as a Mixed Integer Second Order Cone Optimization (MISOCO) problem, for which off-the-shelf optimization solvers as CPLEX or Gurobi are able to solve medium size instances in reasonable CPU time.

Classification Trees (CT) are a family of classification methods based on a
hierarchical relation among a set of nodes. The decision rule  for CT methods is built by recursively partitioning the feature space by means of hyperplanes. At the first stage, a root node for the tree is considered where all the observations belongs to. Branches are sequentially created by splits on the feature space, creating intermediate nodes until a leaf node is reached. Then, the predicted label for an observation is given by the majority class of the leaf node where it belongs to.

\begin{figure}[h]
\centering
\includegraphics[scale=1]{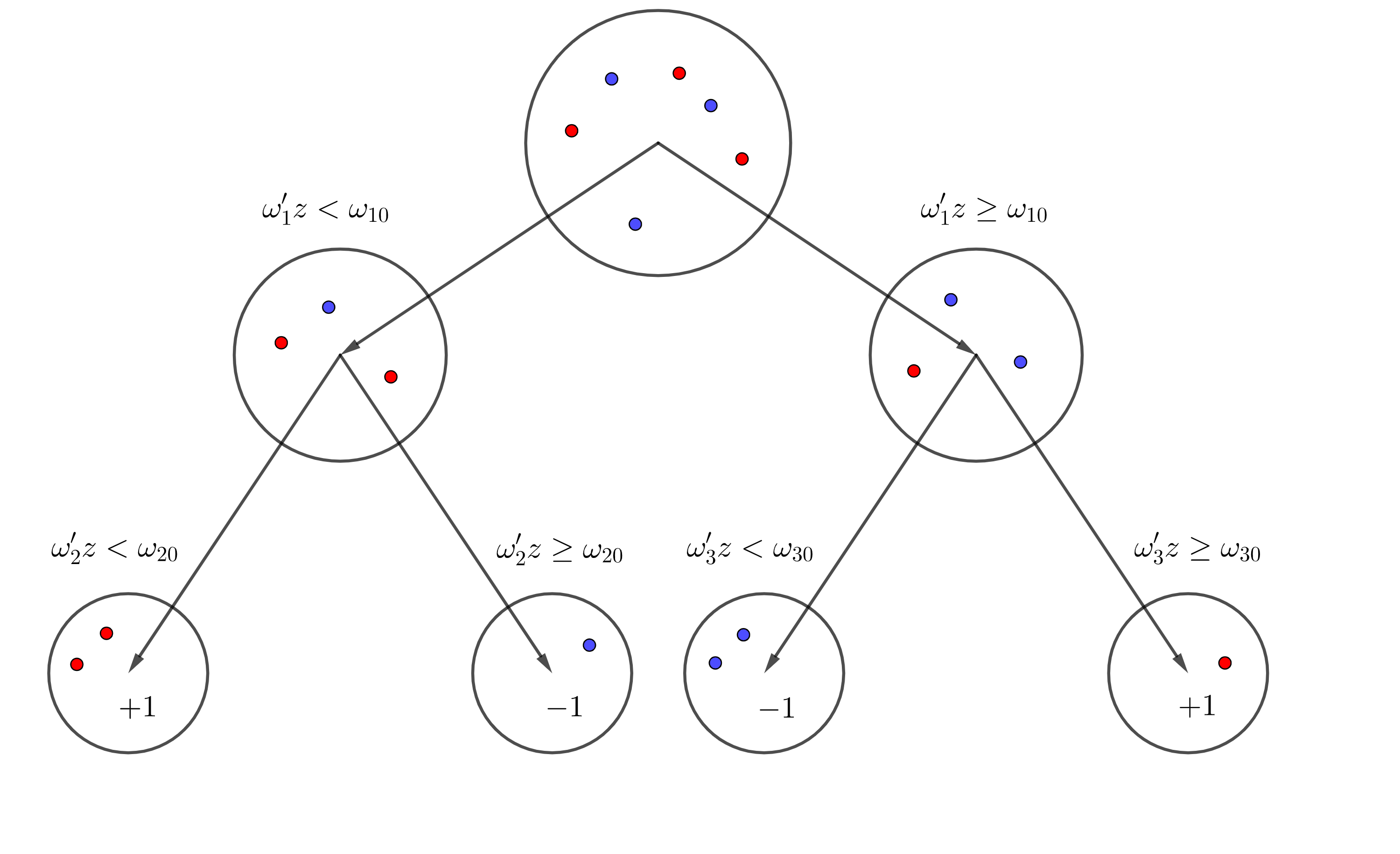}
\caption{Decision tree of depth two.}\label{fig:1}
\end{figure}

Specifically, at each node, $t$, of the tree a hyperplane $\mathcal{H}_t = \{z \in \R^p:  \omega_t' z + \omega_{t 0} = 0\}$ is constructed and the splits are defined as $ \omega_t' z + \omega_{t 0} <0 $ (left branch) and $ \omega_t' z + \omega_{t 0} \geq 0$ (right branch).  In Fig. \ref{fig:1} we show a simple classification tree with depth two, for a small dataset with $6$ observations, that are correctly classified on the leaves.

The most popular method to construct Classification Trees from a training dataset is CART, introduced by Brieman et. al~\cite{cart}. CART is a greedy heuristic approach, which myopically constructs the tree without further foreseen to deeper nodes.  Starting at the root node, it decides the splits by means of hyperplanes minimizing an impurity function in each node. Each split results in two new nodes, and this procedure is repeated until a stopping criterion is reached (maximal depth, minimum number of observations in the same class, etc). CART produces deep trees, which may lead to overfitting in out-of-sample observations. Also, trees are normally subject to a prune process  based on the trade-off between the impurity function reduction and a cost-complexity parameter. The main advantage of CART is that it is easy to implement and fast to train.

On the other hand, Bertsimas and Dunn~\cite{bertsimas2017optimal} have recently proposed an optimal approach to build CTs by solving a mathematical programming problems  which builds the decision tree in a compact model considering its whole structure and at the same time making decisions on pruning or not pruning the branches.

Given a maximum depth, $D$,  for the Classification Tree  it can have at most  $T=2^{D+1} -1$ nodes. These nodes are differentiated in two types:
\begin{itemize}
\item Branch nodes:  $\tau_B = \left\lbrace 1, \ldots , \lfloor T/2\rfloor \right\rbrace$ are the nodes where the splits are applied.
\item Leaf nodes: $\tau_L = \left\lbrace  \lfloor T/2\rfloor ,\ldots, T \right\rbrace $ are the nodes where predictions for observations are performed.
\end{itemize}

We use the following notation concerning the hierarchical structure of a tree:
\begin{itemize}
\item $p(t)$: parent node of node $t$, for $t=1, \ldots, T$.
\item $\tau_{bl}$: set of nodes  that follow the  left branch on the path from their parent nodes. Analogously, we define $\tau_{br}$ as the set of nodes whose right branch has been followed on the path from their parent nodes.
\item $u$: set of nodes that have the same depth inside the tree. We represent by $U$ the whole set of levels. The root node is the zero-level, $u_0$, hence, for a given depth $D$ we have $D+1$ levels, being $u_D$ the set of leaf nodes.
\end{itemize}
OCTs are constructed by minimizing the following objective function:
$$
\displaystyle \sum_{t\in\tau_L}L_t + \alpha \sum_{t\in\tau_B} d_t,
$$
where $L_t$ stands for the misclassification errors at the leaf $t$ (measured as the number of wrongly classified observations in the leaf), and $d_t$ is a binary variable that indicates if a split is produced at $t$. Therefore, the constant $\alpha$ is used to regulate the trade-off between the complexity (depth) and the accuracy (misclassifying errors of the training sample) of the tree. In its simplest version, motivated by what it is done in CART, the splits are defined by means of a single variable, i.e., in the form $\omega_{t j}x_j + \omega_{t 0}\leq 0$. Nevertheless, OCT can be extended to a more complex version where the splits are hyperplanes defined by their normal vector, $a\in\mathbb{R}^p$ which is known as OCT-H. Moreover, a robust version of OCT has also been studied under the noise label scenario in \cite{bertsimas2019robust}.

\section{Optimal Classification Trees with SVM splits and Relabeling (OCTSVM)}\label{sec:3}
This section is devoted to introduce our new classification methodology, namely OCTSVM. The rationale of this approach is to combine the advantage of hierarchical classification methods such as Classification Trees, with the benefits from using distance-based classification errors, by means of hyperplanes maximizing the margin between classes (SVM paradigm). Therefore, this new model rests on the idea of constructing an optimal classification tree in which the splits of the nodes are performed by following the underling ideas of model \eqref{model:1}: 1)  the splits are induced by hyperplanes in which the positive and the negative class are separated maximizing the margin between classes, 2) minimizing the classification errors, and 3) allowing observations to be relabeled along the training process. In contrast to what it is done in other Classification Tree methods, OCTSVM does not make a distinction (beyond the hierarchical one) between branch and leaf nodes, in the sense that RE-SVM based splits are sequentially applied in each node, and the final classification for any observation comes from the hyperplanes resulting at the last level of the tree,  in case there are no pruned branches, or at the last node where a split was made in case of a pruned branch.

\begin{figure}[h]
\begin{center}
\includegraphics[scale=0.7]{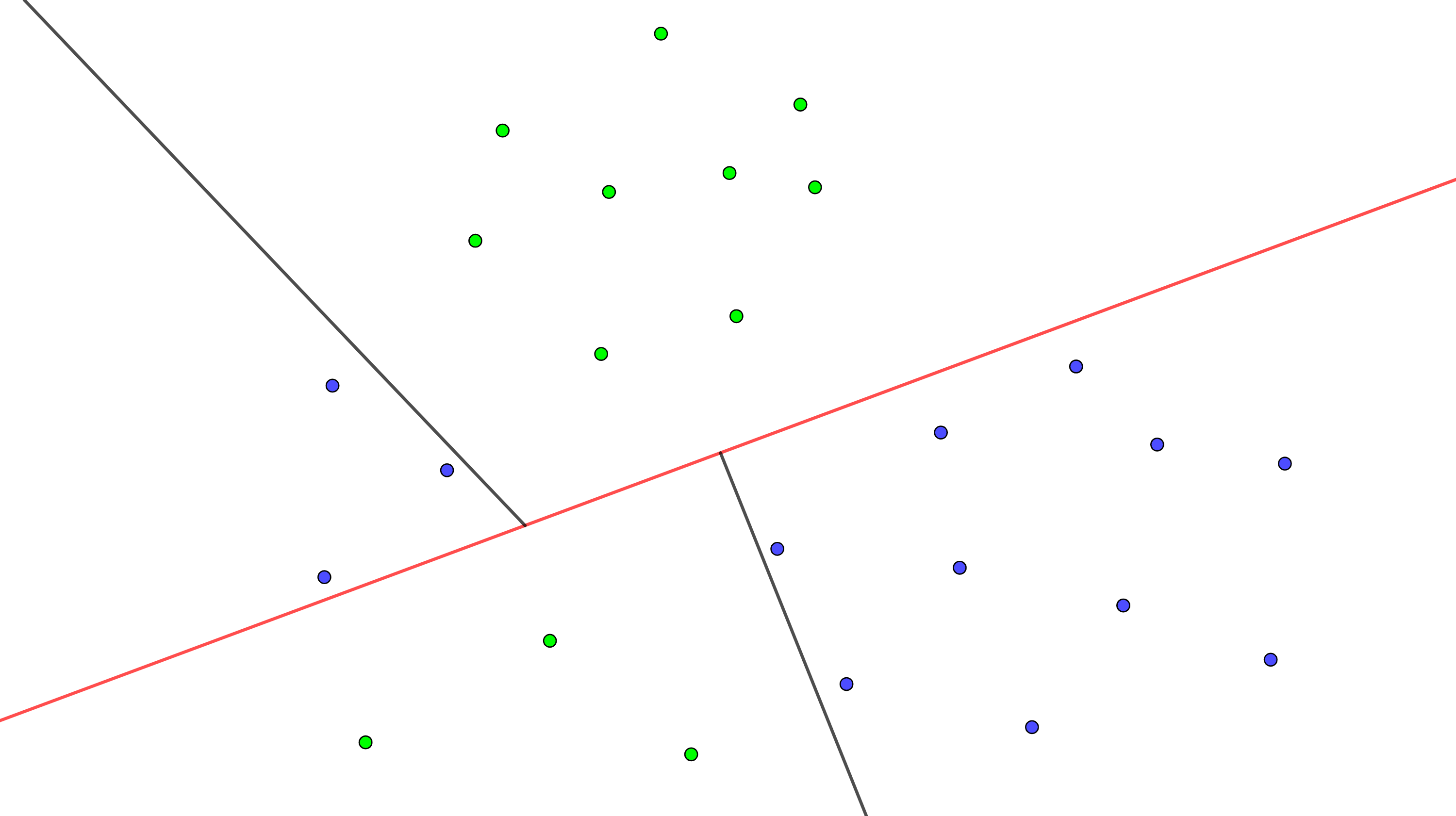}~
\includegraphics[scale=0.7]{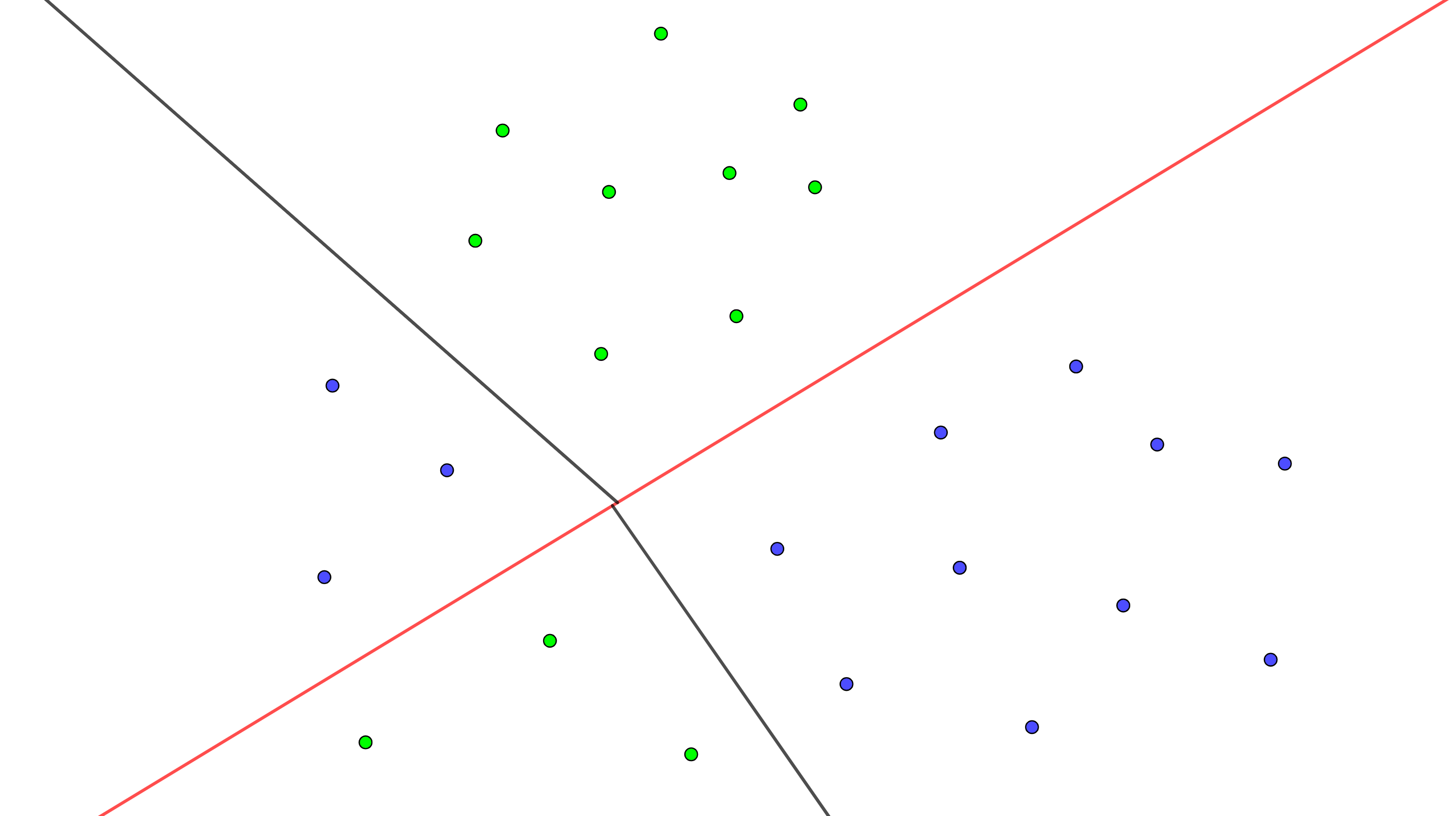}
\end{center}
\caption{Optimal solutions for OCT-H with $D=2$ (left) and OCTSVM with $D=1$ (right).}\label{fig:2}
\end{figure}

As it has been pointed out before, OCT-H is a classification tree that allows the use of general (oblique) hyperplane splits, which is built by solving a single optimization problem that takes into account the whole structure of the tree. Nevertheless, despite the good results this method has proven to obtain in real classification problems, a further improvement is worth to be considered. In Fig. \ref{fig:2} we see a set of points in the plane differentiated by colors in two classes. Looking at the left picture, one can see one of the optimal solutions of OCT-H for depth equal to two, where the red  hyperplane is the split applied at the root node and the black ones are applied on the left and right descendants, which define the four leaves. This solution is optimal, for a certain value of the cost-complexity parameters, since it does not make any mistakes on the final classification. Nevertheless, since this method does not have any kind of control on the distances from points to the hyperplanes, one can observe that the blue class has very tiny margins at the leaves, and hence, for this class, misclassification errors are more likely to occur in out-of-sample observations. On the other hand, on the right side of Fig. \ref{fig:2} one sees another possible optimal solution for the OCTSVM model with depth equal to one. Again, the red hyperplane is the split applied at the root node and the black ones are the classification splits applied at the two leaves. Despite these two methods are obtaining a perfect classification on the training sample, Fig. \ref{fig:2} shows that OCTSVM provides a more balanced solution than OCT-H since it has wider margins between both classes, what could be translated into a higher accuracy for out of sample observations.

In order to formulate the OCTSVM as a MINLP, we will start describing the rationale of its objective function that must account for the margins induced by the splits at all nodes, the classification errors, the penalty paid for relabelling observations and the cost complexity of the final classification tree. To formulate the above concepts we need different sets of variables. First of all, we consider continuous variables: $\omega_t \in \mathbb{R}^p,\  \omega_{t_0} \in \R,\ t = 1,\ldots, T$, which represent the coefficients and the intercept of the hyperplane performing the split at node $t$. Taking into account that the margin of a hyperplane is given by $\frac{2}{||\omega_t||}$, maximizing the minimum margin between classes induced by the splits can be done introducing an auxiliary variable $\delta\in\mathbb{R}$ (that will be minimized in the objective function) which is enforced by the following constraints:

\begin{align}
& \frac{1}{2}||\omega_t||_2 \leq \delta  & \forall t = 1 , \ldots , T. \label{octsvm_1}
\end{align}

Once the maximization of the margin is set, we have to model the minimization of the errors at the nodes, whereas at the same time we minimize the number of relabelled observations. These two tasks are accomplished by the variables $e_{it} \in \mathbb{R},\ i = 1,\ldots ,n, \ t=1,\ldots ,T$, that account for the  misclassification error of observation $i$ at  node $t$, and $\xi_{it}\in\left\lbrace 0 ,1\right\rbrace$ binary variables modelling whether observation $i$ is relabeled or not at node $t$. If $c_1$ and $c_2$ are the unit costs of misclassification and relabelling, respectively, our goal is achieved adding to the objective function the following two terms:
$$\displaystyle c_1\sum_{i=1}^n\sum_{t=1}^T e_{it}+ c_2\sum_{i=1}^n\sum_{t=1}^T \xi_{it}.$$\\
The correct meaning of these two sets of variables must be enforced by some families of constraints that we describe next. Nevertheless, for the sake of readability before describing those constraints modeling these $e_{it}$ and $\xi_{it}$,  we must introduce another family of variables  the $\beta_{it} \in\mathbb{R}^p, \beta_{i0} \in\mathbb{R}, \ i=1,\ldots, n, \ t=1,\ldots, T$, which are continuous variables equal to the coefficients of the separating hyperplane at node $t$ when observation $i$ is relabelled, and equal to zero otherwise. In addition, we consider binary variables $z_{it}\in\left\lbrace 0 , 1 \right\rbrace$ needed to control whether  observation $i$ belongs to node $t$ of the tree. Now, putting all these elements together, as it is done in RE-SVM, we can properly define the splits and their errors at each node of the tree using the following constraints:
\begin{align}
 y_i(\omega_t'x_i + \omega_{t0})-2y_i(\beta_t'x_i + \beta_{it0}) \geq 1 - e_{it} -M(1-z_{it}), \quad \left\{\begin{array}{c} \forall i=1,\ldots,N,\\ t=1,\ldots,T,\end{array}\right.\label{octsvm_2}\\
  \beta_{itj} = \xi_{it}\omega_{tj}, \quad \forall i=1,\ldots,N, t=0,\ldots,T, \ j=0,\ldots,p.\label{octsvm_3}
\end{align}
Constraints \eqref{octsvm_3}, which can be easily linearized, are used to define the $\beta_{it}$ variables: they equal the $\omega_t$ variables when the observation $i$ is relabelled ($\xi_i = 1$), and are equal to zero otherwise ($\xi_i = 0$). On the other hand, constraints \eqref{octsvm_2} control the relabelling at each node of the tree. If an observation $i$ is in node $t$ ($z_{it}=1$), and $\xi_i=0$, we obtain the standard SVM constraints for the separating hyperplane. Nevertheless, if $\xi_i=1$, then the separating constraints are applied for the observation $i$ as if its class were the opposite to its actual one, i.e., as if observation $i$ is relabeled. Moreover, since $M$ is a big enough constant, these constraints do not have any kind of impact in the error variables of node $t$ if observation $i$ does not belong to this node ($z_{it}=0$).\\
The final term to be included in the objective function of the problem is the complexity of the resulting tree. Following the approach in OCT and OCT-H, we consider binary variables $d_t\in\left\lbrace 0 ,1\right\rbrace, \ t=1,\ldots, T$, that account whether a split is applied at node $t$. Thus, to control the tree complexity resulting of the process, we minimize the sum of these variables multiplied by a cost penalty $c_3$. Gathering all the components together, the objective function to be minimized in our problem results in
$$\displaystyle \delta + c_1\sum_{i=1}^n\sum_{t=1}^T e_{it}+ c_2\sum_{i=1}^n\sum_{t=1}^T \xi_{it} + c_3\sum_{t=1}^T d_t.$$
According to this, we need to consider the following constraints so as to obtain a suitable modeling for the $d_t$ variables:
\begin{align}
& \|\omega_t \|_2 \leq Md_t, &\forall t=1,\ldots,T.\label{octsvm_4}
\end{align}
By imposing this, and being $M$ a big enough constant, $d_t = 0$ only if $\|\omega_t\|_2=0$, that is, if there is no split at node $t$. On the other hand, in case a split is applied at node $t$, $\|\omega_t\|_2>0$, and hence $d_t$ is forced to assume the value one. Another point that is important to remark about the $d_t$ variables is that once a non-leaf node does not split (that is, the corresponding branch is pruned at this node), the successors of node $t$ can not make splits either to maintain the correct hierarchical structure of the tree. Recalling that $p(t)$ is the parent node of node $t$, we can guarantee this relationship throughout the following constraints
\begin{align}
& d_t \leq d_{p(t)}, &\forall t=1,\ldots,T.\label{octsvm_5}
\end{align}
There are still some details left that must be imposed to make the model work as required. In the decision tree, observations start at the root node and they advance descending through the levels of the tree until they reach a leaf or a pruned node. Hence, we have to guarantee that observations must belong to one, and only one, node per level. By means of the $z_{it}$ variables, this can be easily done by the usual assignment constraints applied in each level, $u\in U$, of the tree:
\begin{align}
& \sum_{t\in u}z_{it} = 1, &\forall i=1,\ldots,N,\ u\in U.\label{octsvm_6}
\end{align}
Moreover, for consistency in the relation between a node and its ancestor, it is clear that if observation $i$ is in node $t$ ($z_{it}=1$), then, observation $i$ must be also in the parent of node $t$ ($z_{ip(t)}=1$), with the only exception of the root node. Besides, if observation $i$ is not in node $t$ ($z_{it}=0$), then $i$ can not be in its successors, and this is modeled by adding the following constraints to the problem:
\begin{align}
& z_{it} \leq z_{ip(t)}, &\forall i=1,\ldots,N, t =2,\ldots, T.\label{octsvm_7}
\end{align}
So far, the OCTSVM model has everything it needs to properly perform the splits by following the RE-SVM rationale described in \eqref{model:1}, taking into consideration the tree complexity, and maintaining the hierarchical relationship among nodes. The last element that we need to take care of, to assure the correct performance of the whole model, is to define how observations follow their paths inside the tree. We get from constraints \eqref{octsvm_7} that observations move from parent to children (nodes), but every non terminal node has a left and a right child node, and we need to establish how observations take the left or the right branch. Since the splits are made by the separating hyperplane, we force observations that lie on the positive half space of a hyperplane to follow the right branch of the parent node, and observations that lie con the negative one to take the left branch. This behavior is modeled with the binary variables $\theta_{it}\in\left\lbrace 0,1 \right\rbrace$, that are used to identify if observation $i$ lies in the positive half space of the separating hyperplane at node $t$, $\theta_{it}=1$, or if observation $i$ lies on the negative half space, $\theta_{it}=0$. By considering $M$ a big enough constant, the correct behavior of path followed by the observations is enforced by the following constraints:
\begin{align}
& \omega_t'x_i + \omega_{t0} \geq -M(1-\theta_{it}), &\forall i=1,\ldots,N, t =1,\ldots, T,\label{octsvm_8}\\
& \omega_t'x_i + \omega_{t0} \leq M\theta_{it}, &\forall i=1,\ldots,N, t =1,\ldots, T.\label{octsvm_9}
\end{align}
Hence, by making use of these $\theta_{it}$ variables, and distinguishing between nodes that come from left splits, $\tau_{bl}$ (nodes indexed by even numbers), and right splits, $\tau_{br}$ (nodes indexed by odd numbers), we control that the observations  follow the paths through the branches in the way we described above throughout the following constraints:
\begin{align}
& z_{ip(t)} - z_{it} \leq \theta_{ip(t)}, & \forall i=1,\ldots,N, t \in \tau_{bl} \label{octsvm_10}\\
& z_{ip(t)} - z_{it} \leq 1-\theta_{ip(t)}, & \forall i=1,\ldots,N, t \in \tau_{br} \label{octsvm_11}
\end{align}
According to constraints \eqref{octsvm_10}, if an observation $i$ is on the parent node of an even node $t$ ($z_{ip(t)}=1$), and $i$ lies on the negative half space of the hyperplane defining the split on $p(t)$ ($\theta_{ip(t)}=0$), then $z_{it}$ is forced to be 1. Hence, $\theta_{ip(t)}=0$ implies that observation $i$ takes the left branch to the child node $t \in \tau_{bl}$. Moreover, we can see that this  constraint is consistent since if $z_{ip(t)}=1$, but observation $i$ is not in the left child node, $z_{it}=0$, $t\in\tau_{bl}$, then $\theta_{ip(t)}$ equals 1, and  which means that observation $i$ lies on the positive half space of the hyperplane of $p(t)$. On the other hand, constraints \ref{octsvm_11} are similar but for the right child nodes, $\tau_{br}$. If an observation $i$ is in the parent node of an odd node $t \in \tau_{br}$, and $i$ lies on the positive half space of the hyperplane of $p(t)$ ($\theta_{ip(t)} = 1$), then, $z_{it}=1$ what means that observation $i$ has to be on node $t$.\\

Gathering all the constraints together, the OCTSVM model is obtained by solving the following MINLP:

\begin{align} \displaystyle  \min \ & \delta + c_1\sum_{i=1}^n\sum_{t=1}^T e_{it}+ c_2\sum_{i=1}^n\sum_{t=1}^T \xi_{it} + c_3\sum_{t=1}^T d_t
\tag{${\rm OCTSVM}$} & \label{model_1}\\
\mbox{s.t. } & \frac{1}{2}||\omega_t||_2 \leq \delta,  \quad \forall t = 1 , \ldots , T, \nonumber\\
& y_i(\omega_t'x_i + \omega_{t0})-2y_i(\beta_t'x_i + \beta_{it0}) \geq 1 - e_{it} -M(1-z_{it}), \qquad \forall i=1,\ldots,N, t=1,\ldots,T,\nonumber\\
&  \beta_{itj} = \xi_{it}\omega_{tj}, \qquad \forall i=1,\ldots,N, t=0,\ldots,T, \ j=0,\ldots,p,\nonumber\\
& \|\omega_t \|_2 \leq Md_t,  \qquad \forall t=1,\ldots,T,\nonumber\\
& d_t \leq d_{p(t)} , \qquad \forall t=1,\ldots,T,\nonumber\\
& \sum_{t\in u}z_{it} = 1,  \qquad \forall i=1,\ldots,N,\ u\in U,\nonumber\\
& z_{it} \leq z_{ip(t)} , \qquad \forall i=1,\ldots,N, t =2,\ldots,T,\nonumber\\
& \omega_t'x_i + \omega_{t0} \geq -M(1-\theta_{it}), \qquad \forall i=1,\ldots,N, t =1,\ldots, T,\nonumber\\
& \omega_t'x_i + \omega_{t0} \leq M\theta_{it},  \qquad \forall i=1,\ldots,N, t =1,\ldots, T,\nonumber\\
& z_{ip(t)} - z_{it} \leq \theta_{ip(t)} , \qquad  \forall i=1,\ldots,N, t \in \tau_{bl}, \nonumber\\
& z_{ip(t)} - z_{it} \leq 1-\theta_{ip(t)},  \qquad \forall i=1,\ldots,N, t \in \tau_{br}, \nonumber\\
& e_{it} \in\mathbb{R}^+, \beta_{it}\in\mathbb{R}^p,\beta_{it0}\in\mathbb{R},  \xi_{it},z_{it}, \theta_{it} \in\left\lbrace 0,1 \right\rbrace, \,  \forall i=1,\ldots,N, t=1,\ldots,T,\nonumber\\
&\omega_t\in\mathbb{R}^p, \omega_{t0} \in\mathbb{R}, d_t \in\left\lbrace 0,1 \right\rbrace, \forall t=1,\ldots,T.\nonumber
\end{align}

\section{Experiments}\label{sec:4}

In this section we present the results of our computational experiments. Four different Classification Tree-based methods are compared, CART, OCT, OCT-H and OCTSVM,  on eight popular real-life datasets from UCI Machine Learning Repository~\cite{UCI}. The considered datasets together with their dimensions ($n$: number of observations and $p$: number of features) are shown in Table \ref{table:1}.

\begin{table}[h]
\centering
\begin{tabular}{l|cc}
{\bf Dataset} & $n$ & $p$ \\
\hline\hline
\texttt{Australian}  & 690 & 14 \\
\texttt{BreastCancer}  & 683 & 9 \\
\texttt{Heart}  & 270 & 13 \\
\texttt{Ionosphere}  & 351 & 34 \\
\texttt{MONK's}  & 415 & 7 \\
\texttt{Parkinson}  & 240 & 40 \\
\texttt{Sonar}  & 208 & 60 \\
\texttt{Wholesale}& 440 & 7 \\
\hline
\end{tabular}
\caption{Datasets used in our computational experiments.\label{table:1}}
\end{table}
Our computational experience focuses on the analysis of the accuracy of the different classification trees-based methods. This analysis is based in four different experiments per each one of the eight considered dataset. Our goal is to analyze the robustness of the different methods against label noise in the data. Therefore, in our experiments we use, apart from the original datasets, three different modifications where in each one of them a percentage of the labels in the sample are randomly flipped. The percentages of flipped labels range in $\left\lbrace 0\%, 20\%, 30\%, 40\% \right\rbrace$, where $0\%$ indicates that the original training data set is used to construct the tree.

We perform a 4-fold cross-validation scheme, i.e., datasets are split into four random train-test partitions. One of the folds is used for training the model while the rest are used for testing. When testing the classification rules, we compute the accuracy, in percentage, on out of sample data:
\begin{center}
$\textit{ACC} = \frac{\#\textit{Well Classified Test Observations}}{\#\textit{Test Observations}}\cdot 100$.
\end{center}
The CART method was coded in \texttt{R} using the \texttt{rpart} library. OCT, OCT-H and OCTSVM mathematical programming models were coded in \texttt{Python} and solved using \texttt{Gurobi} 8.1.1 on a PC Intel Core i7-7700 processor at 2.81 GHz and 16GB of RAM. A time limit of 30 seconds was set for training.

The calibration of the parameters of the different optimization-based models compared in these experiments was set as follows:
\begin{itemize}
\item For OCTSVM we used a grid on $\left\lbrace 10^i: i=-5,\ldots , 5 \right\rbrace$ for the constants $c_1$ and $c_2$, and a smaller one $\left\lbrace 10^i: i=-2,\ldots , 2 \right\rbrace$ for $c_3$.
\item For CART, OCT and OCT-H we used the same grid $\left\lbrace 10^i: i=-5,\ldots , 5 \right\rbrace$ for the cost-complexity constant, and a threshold on the minimum number of observations per leaf equal to a 5$\%$ of the training sample size.
\end{itemize}
Last to mention, the depth, $D$, considered in these experiments was equal to three for CART, OCT and OCT-H, whereas for OCTSVM we fixed depth equal to two, creating consequently trees with 3 levels, to set a fair comparison among the different methods.

For each dataset, we replicate each experiment four times. In Table \ref{table:2} we show the average accuracies for each one of the methods and datasets. The best average accuracies are highlighted (boldfaced).

\begin{table}[h]
\begin{center}
\begin{tabular}{|>{\ttfamily}c|c|r|r|r|r|}
\cline{2-6}
\multicolumn{1}{c|}{} & \% Flipped & CART & OCT & OCT-H & OCTSVM \\\hline
\multirow{5}{*}{Australian} & 0 & 82.99 & 85.44 & 85.16 & \textbf{86.34} \\
& 20 & 74.85 & 83.45 & 79.88 & \textbf{84.55} \\
& 30 & 66.87 & 79.15 & 71.55 & \textbf{80.24} \\
& 40 & 56.93 & 71.34 & 64.25 & \textbf{73.89} \\\cline{2-6}
& \multicolumn{1}{r|}{Average} & 70.41 & 79.85 & 75.21 & \textbf{81.26} \\
\hline
\multirow{5}{*}{BreastCancer} & 0 & 92.22 & 93.18 & 94.21 & \textbf{96.25} \\
& 20 & 90.47 & 90.92 & 89.38 & \textbf{91.57} \\
& 30 & 83.29 & \textbf{90.87} & 84.83 & 87.98 \\
& 40 & 77.75 & \textbf{86.50} & 76.35 & 81.92 \\\cline{2-6}
& \multicolumn{1}{r|}{Average} & 85.93 & \textbf{90.37} & 86.19 & 89.43 \\
\hline
\multirow{5}{*}{Heart} & 0 & 73.66 & 73.88 & 78.86 & \textbf{84.13} \\
& 20 & 72.98 & 71.63 & 74.22 & \textbf{82.61} \\
& 30 & 68.22 & 70.18 & 70.88 & \textbf{80.51} \\
& 40 & 62.36 & 64.52 & 65.30 & \textbf{76.19} \\\cline{2-6}
& \multicolumn{1}{r|}{Average} & 69.31 & 70.05 & 72.32 & \textbf{80.86} \\
\hline
\multirow{5}{*}{Ionosphere} & 0 & 83.08 & 81.80 & 84.40 & \textbf{85.51} \\
& 20 & 75.65 & 79.59 & 75.08 & \textbf{80.51} \\
& 30 & 70.02 & 75.20 & 71.93 & \textbf{77.79} \\
& 40 & 60.17 & 70.65 & 65.87 & \textbf{75.71} \\\cline{2-6}
& \multicolumn{1}{r|}{Average} & 72.23 & 76.81 & 74.32 & \textbf{79.88} \\
\hline
\multirow{5}{*}{MONK's} & 0 & 60.67 & 59.03 & \textbf{61.63} & 61.26 \\
& 20 & 57.11 & 59.59 & 59.33 & \textbf{60.48} \\
& 30 & 57.19 & 57.87 & 59.13 & \textbf{60.09} \\
& 40 & 54.01 & 55.25 & 57.01 & \textbf{60.12} \\\cline{2-6}
& \multicolumn{1}{r|}{Average} & 57.25 & 57.94 & 59.28 & \textbf{60.49} \\
\hline
\multirow{5}{*}{Parkinson} & 0 & 74.73 & 74.38 & 74.69 & \textbf{81.52} \\
& 20 & 67.03 & 70.82 & 66.66 & \textbf{76.24} \\
& 30 & 63.68 & 71.39 & 62.81 & \textbf{73.01} \\
& 40 & 56.79 & 59.72 & 58.46 & \textbf{68.73} \\\cline{2-6}
& \multicolumn{1}{r|}{Average} & 65.56 & 69.08 & 65.66 & \textbf{74.88} \\
\hline
\multirow{5}{*}{Sonar} & 0 & 65.36 & 66.90 & 71.63 & \textbf{74.76} \\
& 20 & 57.41 & 60.38 & 65.45 & \textbf{69.92} \\
& 30 & 54.64 & 58.97 & 62.53 & \textbf{67.30} \\
& 40 & 55.16 & 57.77 & 60.33 & \textbf{63.84} \\\cline{2-6}
& \multicolumn{1}{r|}{Average} & 58.14 & 61.01 & 64.99 & \textbf{68.96} \\
\hline
\multirow{5}{*}{Wholesale} & 0 & 90.28 & 90.01 & \textbf{90.51} & 89.75 \\
& 20 & 83.79 & \textbf{87.84} & 85.43 & 82.06 \\
& 30 & 78.54 & \textbf{84.58} & 77.91 & 78.13 \\
& 40 & 69.52 & \textbf{75.79} & 72.56 & 71.72 \\\cline{2-6}
& \multicolumn{1}{r|}{Average} & 80.53 & \textbf{84.56} & 81.60 & 80.42 \\
\hline
\multicolumn{2}{|c|}{Total Averages} & 69.92 & 73.71 & 72.44 & \textbf{77.02} \\
\hline
\end{tabular}%
\caption{Averaged accuracy results of the computational experiments.}\label{table:2}
\end{center}
\end{table}

Observe that when $0\%$ of the  training labels are flipped (i.e., the original dataset), a general trend in terms of accuracy can be observed: \textbf{CART $<$ OCT $<$ OCT-H $<$ OCTSVM}, with the only exception of \texttt{Wholesale} in which CART obtains a non-meaningful slightly larger average accuracy. We would like to emphasize  that these results are not surprising. Indeed, on the one hand, OCT is an optimal classification version of the CART algorithm, and hence better results should be expected from this model, as already shown in the computational experience in \cite{bertsimas2017optimal}, and also evidenced in our experiments. Moreover, OCT is just a restricted version of OCT-H, in that in the latter, the set of admissible hyperplanes is larger than in the former. Also, as already pointed out in Fig. \ref{fig:2}, OCTSVM goes one step further and, apart from allowing oblique hyperplanes based trees, has a more solid structure due to the margin maximization, the distance based errors and the possibility of relabeling points. All together turns into higher accuracy results as pointed out in our experiments. In some datasets the comparison of the different methods gives rather similar results, however in some others OCTSVM is above a 5$\%$ percent of accuracy with respect to the other three methods.

Turning to the results on datasets with flipped labels on the training dataset, OCTSVM clearly outperforms the rest of methods and consistently keeps its advantage in terms of accuracy with respect to the other three methods. OCT and OCT-H, which are both above CART, alternate higher-lower results among the different datasets. Our  method, namely OCTSVM, clearly is capable to capture the wrongly labeled observations and constructs  classification rules able to reach higher accuracies, even when $40\%$ of the labels were flipped, while other methods, give rise to classifiers that significantly worsen their performance in terms of accuracy in the test sample.

The global average accuracy in  the case that labels are not flipped using OCTSVM is 82.44\%, being 2\% better than OCT-H, 3\% better than OCT and 10\% better than CART. If 40\% of the labels are flipped the situation is similar, being OCTSVM also 10\% better than CART, and then, clearly showing the robustness of our method under label noises.

\section{Conclusions and Further Research}\label{sec:5}

Supervised classification is a fruitful field that has attracted the attention of researchers for many years. One of the methods that has experienced a more in depth transformation in the last years is classification trees. Since the pioneer contribution by Breiman et al. \cite{cart}, where CART was proposed, this technique has included tools from mathematical optimization giving rise to the so called OCT \cite{bertsimas2017optimal,bertsimas2019robust}, methods that are \textit{optimal} in some sense. In spite of that, there is still some more room for further improvements, in particular making classifiers more robust against perturbed datasets. Our contribution is this paper goes in that direction and it augments the catalogue of classification tree methods able to handle noise in the labels of the dataset.

We have proposed a new optimal classification tree approach able to handle label noise in the data. Two main elements support our approach: the splitting rules for the classification trees are designed to maximize the separation margin between classes and wrong labels of the training sample are detected and changed at the same time that the classifier is built. The method is encoded on a Mixed Integer Second Order Order Cone Optimization problem so that one can solve medium size instances by any of the nowadays available off-the-shelf solvers. We report intensive computational experiments on a battery of datasets taken from UCI Machine Learning repository showing the effectiveness and usefulness of our approach.

Future research lines on this topic include the analysis of nonlinear splits when branching  in  OCTSVM, both using kernel tools derived from SVM classifiers or  specific families of nonlinear separators. This approach will result into more flexible classifiers able to capture the nonlinear trends of many real-life datasets. Additionally, we also plan to address the design of math-heuristic algorithms which keep the essence of OCTs but capable to train larger datasets.

%

%

\end{document}